\documentclass[letterpaper, 10 pt, conference]{ieeeconf}  

\IEEEoverridecommandlockouts                              

\usepackage[ruled]{algorithm2e}
\usepackage{algpseudocode}
\usepackage{graphics}
\usepackage{amsmath}
\DeclareMathOperator*{\argmax}{argmax}

\usepackage{amssymb}
\usepackage{epsfig} 
\usepackage{booktabs}
\usepackage{float}
\usepackage{url}
\usepackage{gensymb}
\usepackage{hyperref}

\newcount\Comments 
\usepackage{array,multirow,graphicx}
\usepackage{color}
\usepackage{wrapfig}
\definecolor{darkgreen}{rgb}{0,0.5,0}
\definecolor{purple}{rgb}{1,0,1}
\newcommand{\kibitz}[2]{\ifnum\Comments=1\textcolor{#1}{#2}\fi}

\title{\LARGE \bf
TANDEM3D: Active Tactile Exploration for 3D Object Recognition
}

\author{Jingxi Xu$^{*1}$, Han Lin$^{*1}$, Shuran Song$^1$, and Matei Ciocarlie$^{2}$
\thanks{*Equal contribution.}
\thanks{$^{1}$Department of Computer Science, Columbia University, New York, NY 10027, USA. {\tt\small \{jxu, shurans\}@cs.columbia.edu}, {\tt\small hl3199@columbia.edu}}%
\thanks{$^{2}$Department of Mechanical Engineering, Columbia University, New York, NY 10027, USA. {\tt\small matei.ciocarlie@columbia.edu}}%
\thanks{This work was supported in part by the NSF under grant CMMI-2037101.}
}

\begin{document}

\maketitle
\thispagestyle{empty}
\pagestyle{empty}

\begin{abstract}
Tactile recognition of 3D objects remains a challenging task. Compared to 2D shapes, the complex geometry of 3D surfaces requires richer tactile signals, more dexterous actions, and more advanced encoding techniques. In this work, we propose TANDEM3D, a method that applies a co-training framework for exploration and decision making to 3D object recognition with tactile signals. Starting with our previous work, which introduced a co-training paradigm for 2D recognition problems, we introduce a number of advances that enable us to scale up to 3D. TANDEM3D is based on a novel encoder that builds 3D object representation from contact positions and normals using PointNet++. Furthermore, by enabling 6DOF movement, TANDEM3D explores and collects discriminative touch information with high efficiency. Our method is trained entirely in simulation and validated with real-world experiments. Compared to state-of-the-art baselines, TANDEM3D achieves higher accuracy and a lower number of actions in recognizing 3D objects and is also shown to be more robust to different types and amounts of sensor noise. 
\end{abstract}
\section{Introduction}


Inspired by human's ability to complete complex manipulation tasks in the absence of vision, such as identifying and retrieving small objects from the pockets, tactile sensing is receiving an increasing amount of attention in the robotics research community, especially when vision is unavailable due to occlusions, lighting, etc. However, touch sensing is fundamentally an active modality and tactile signals can be expensive to gather without smart guidance. 



The challenges with tactile sensors are further exacerbated when interacting with 3D objects. Many heuristics that work well for exploring 2D objects are not well-defined in 3D. For example, contour following is an efficient exploration strategy in 2D space, but at any point on a 3D surface, there are infinite possible contours to follow. Furthermore, simple object representation is affected by the curse of dimensionality: while occupancy grids are effective representations for 2D shapes, their 3D equivalent, voxel grids, quickly become intractable to use except for very low resolutions.


In previous work, we introduced TANDEM~\cite{xu2022tandem}, a method to co-learn an exploration policy with decision making for recognizing 2D objects. While TANDEM is shown to outperform state-of-the-art baselines at its intended application, it does not scale to 3D problems. Firstly, TANDEM uses binary collision signals, rearranges them into an occupancy grid, and then encodes the grid with convolutional neural networks (CNNs). The direct extension to 3D would be voxel grids and 3D convolution; however, the size of voxel grids grows cubically with the size of the workspace, which is not memory efficient. Secondly, the tactile sensor moves in a 2D horizontal plane, limiting the ability of the robotic manipulators to move flexibly and freely on a 3D surface.

\begin{figure}[t]
    \centering
    \includegraphics[width=\linewidth]{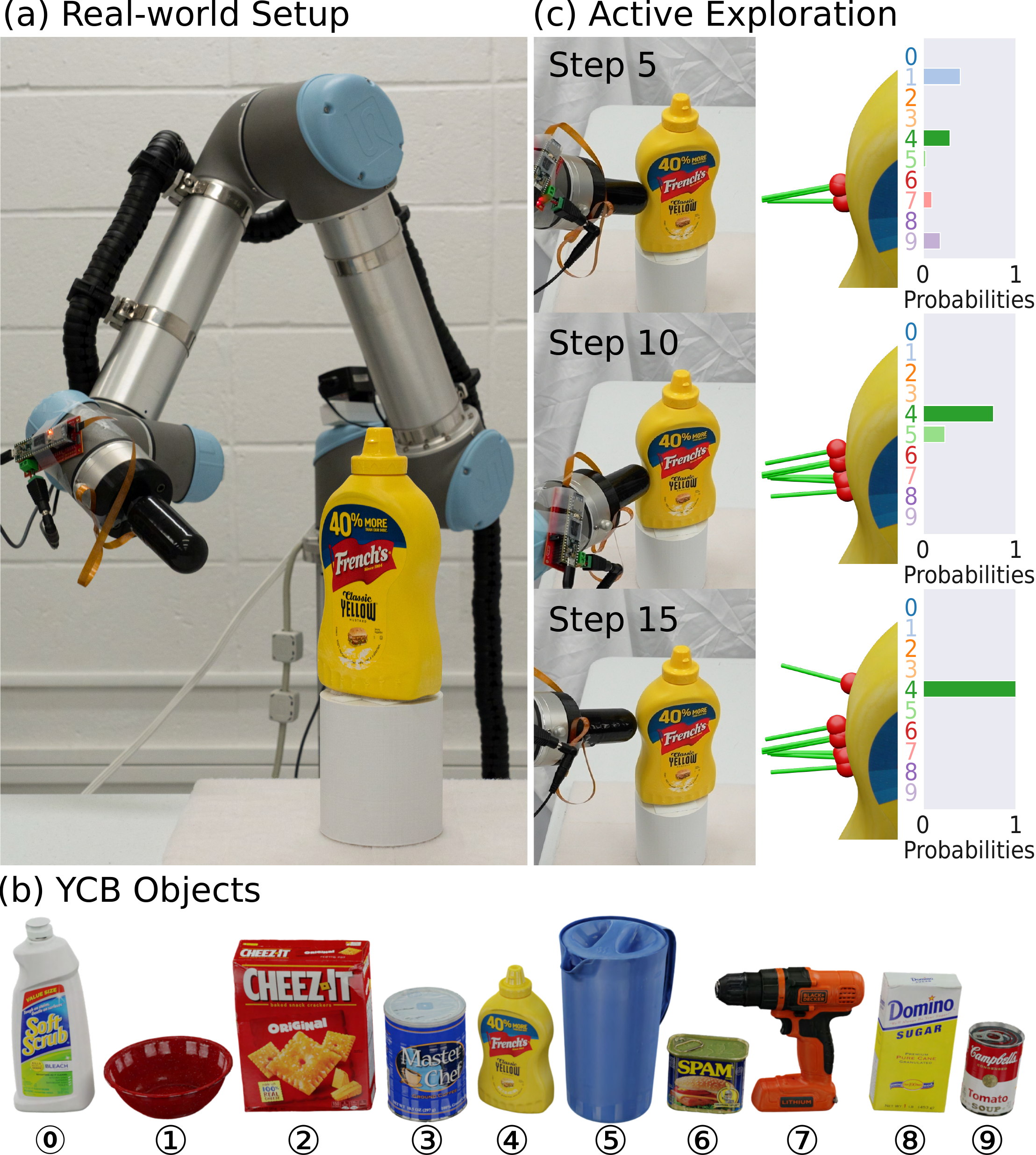}
    \caption{Tactile recognition of 3D objects. (a) Real-world setup. Our tactile finger is mounted on a robot arm. The object is placed on a flat surface with its Z-axis facing up, but the exact position on the horizontal plane and the orientation around the vertical axis are all unknown. (b) A known set of 10 YCB objects, selected to cover a variety of concave and convex 3D shapes and sizes. (c) Using our method, the robot actively collects data and quickly recognizes this object (mustard bottle). It first moves down the vertical edge and makes an initial hypothesis between objects 4 (mustard bottle) and 5 (pitcher base). It then moves up to obtain a final contact on a distinguishable geometry before making the correct identification.}
    \label{fig:teaser}
    \vspace{-0.2in}
\end{figure}

In this work, we focus on learning an active exploration policy for the tactile recognition of 3D objects. We show that a co-training approach can be used for 3D object recognition, where it outperforms state-of-the-art alternatives. To achieve this performance level, we rely on the following advances: (1) We design our policies to make use of a richer action space by enabling 6 degrees of freedom (DOF) movements of the tactile sensor. With 6DOF movements, critical discriminative points can be obtained through small angle adjustments, especially for our multi-curved tactile finger with all-around sensing coverage. (2) We use a richer representation of contact, going beyond binary cell occupancy. In particular, we use contact locations and surface normals, which provide important information about 3D shapes while still being simple enough to simulate, which enables a large amount of training in simulation and zero-shot transfer onto real robots. (3) We propose a more advanced, learning-based encoding of the tactile sensor data. We store tactile points into an unordered point set and encode them with PointNet++~\cite{qi2017pointnet++}. We show that this method provides effective representation for discrimination and exploration, and allows us to scale from binary 2D grids to 3D shape representation.
 
To the best of our knowledge, we are the first to co-train a 6DOF exploration and decision-making strategy for 3D object tactile recognition, a method that we dub TANDEM3D. We show that TANDEM3D outperforms a variety of baselines for recognizing a set of 3D objects placed under partially unknown positions and orientations. Our baselines include a learning-based all-in-one policy that does not distinguish between discriminator and explorer, heuristic-based exploration policies (such as info-gain, contour following, etc.), and an ICP discrimination policy. Compared to these methods, TANDEM3D recognizes objects with a higher success rate and is also more robust to different types and amounts of sensor noise.  We also validate our method on real robot experiments using a subset of the YCB objects~\cite{calli2017yale}.

\section{Related Work}

\subsection{3D Object Recognition with Tactile Sensors}
Object recognition is the problem of identifying one out of a set of known objects. Traditionally, vision has been the predominant sensing modality for such tasks but due to its sensitivity to lighting, occlusions, etc., object recognition with only tactile signals has received increasing attention.

While there are many works on tactile recognition of 2D objects~\cite{xu2022tandem, martinez2013active, yu2015shape, suresh2020tactile, liu2017recent}, tactile recognition of 3D objects is less addressed in the community. \cite{schmitz2014tactile} uses power grasps to collect tactile data and then train a deep neural network for recognition. However, there is no active exploration of the objects and instead, the object is handed to the robot in random poses. Both \cite{schneider2009object} and \cite{pezzementi2011tactile} use bag-of-features for their identification strategy. While \cite{schneider2009object} collects data with predefined repetitive grasps at different heights, \cite{pezzementi2011tactile} develops a surface-contact-control scheme to explore the object after the first contact is made. To cope with the unknown object movement, \cite{luo2015novel} proposes a new tactile-SIFT descriptor but the contact is made following a tightly-controlled prescribed heuristic. 

\subsection{Tactile Exploration Policy}
In this paper, we group exploration policies into heuristic-based and learning-based categories. Contour-following is one the of most popular exploration heuristics used for 2D objects~\cite{martinez2013active, yu2015shape, suresh2020tactile}; however, contour-following in 3D is not well-defined. Other heuristics include information gain (uncertainty reduction)~\cite{hebert2013next, xu2013tactile, schneider2009object, driess2017active}, attention cubes~\cite{rajeswar2021touch}, Monte Carlo tree search~\cite{zhang2017active} and dynamic potential fields~\cite{bierbaum2009grasp}. Other works also study the interplay between decision making and exploration where they pre-train a discriminator with a pre-collected dataset and then use it to estimate action quality with Bayesian methods to reduce uncertainty~\cite{fishel2012bayesian, lepora2013active, martinez2017active, kaboli2017tactile, kaboli2019tactile}. Depending on the task, heuristic-based exploration policies can improve efficiency and require no training, but their performance can deteriorate significantly when sensor noise shows up.

In contrast, learning-based exploration policies can be trained in an unsupervised fashion through trial and error. Noise can be incorporated into the training process so that the policy can be more robust~\cite{xu2022tandem}. However, \cite{xu2022tandem} only handles 2D objects and their method does not scale directly to 3D objects. Our sensor moves in 6DOF action space and our encoder builds 3D object representation from contact positions and normals using PointNet++. Compared to works that rely on high-dimensional or multimodal tactile data (such as~\cite{fishel2012bayesian, lepora2013active, martinez2017active}), contact locations and normals are easy to simulate which enables a large amount of training in simulation and zero-shot transfer to real-world experiments. 
\section{Method}

\begin{figure*}[t!]
\centering
\includegraphics[width=0.95\textwidth]{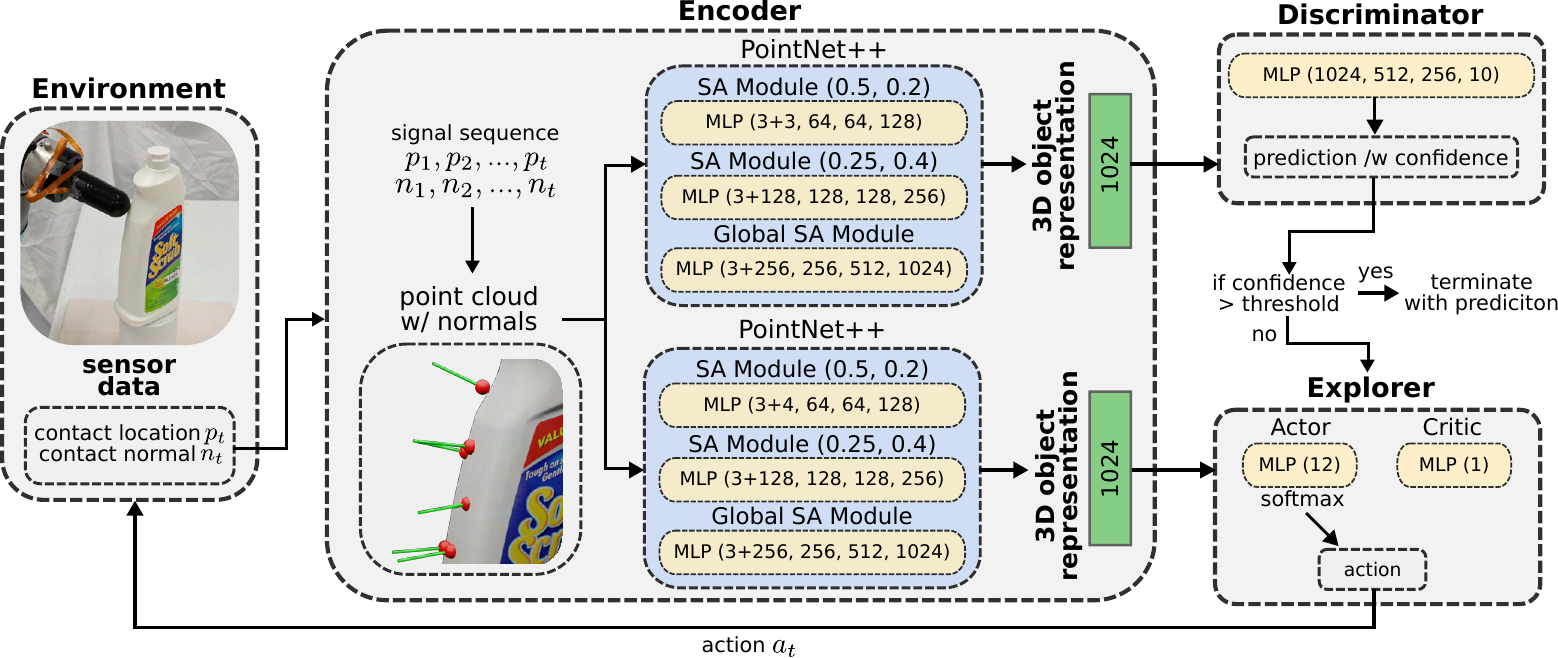}
\caption{An overview of our co-training architecture. Our proposed encoder encodes a sequence of tactile signals into 3D object representation using PointNet++. The parameters of the set abstraction (SA) module are the sampling ratio of the sampling layer and the grouping radius of the grouping layer. The multilayer perceptron (MLP) networks are highlighted in yellow rectangles and their parameters are the number of nodes in each fully-connected layer.}
\label{fig:overview}
\vspace{-0.15in}
\end{figure*}

Our method builds on the co-training framework that we previously introduced and validated for 2D environments~\cite{xu2022tandem}. Referred to as TANDEM, it jointly learns exploration and decision making: an exploration policy (i.e., \textit{explorer}) determines the next action to collect more information, while a decision-making component (i.e., \textit{discriminator}) outputs a prediction for object identity along with a confidence value. If the confidence is high enough, it terminates the exploration. A third component, the \textit{encoder}, encodes the sparse and local tactile signals into a global object representation that is used by the explorer and discriminator. 

While TANDEM is effective for recognizing 2D polygons, it does not scale to 3D problems. Reducing tactile data to occupancy grids discards valuable information that could be used to discriminate between complex surfaces. Furthermore, the grid representation increases exponentially in size with the number of dimensions, which prevents learning. Finally, the limited action set is unable to quickly collect discriminative data on 3D geometry. We address these aspects below.

\subsection{6DOF Action Space}
\label{action}
In order to efficiently explore the complex geometry of 3D objects, we allow the finger to move in 6DOF action space $\mathcal{A}$. Using the top of the fingertip hemisphere as the reference point, we discretize the 6DOF action into small translation (x, y, z) and rotation (roll, pitch, yaw) steps, all with respect to the workspace frame. The robot picks one of x, y, z, roll, pitch, yaw, and can either increase or decrease for a small step (1cm for translation and 15 degrees for rotation). Thus, there are $2 \times 6 = 12$ actions in total. Within these small movements, the tactile finger constantly checks collision and produces contact locations and normals when  collision.

\subsection{3D Object Representation}
TANDEM uses binary collision signals and consequently rearranges them into an occupancy grid. They encode occupancy grids with CNNs. A straightforward extension into 3D space is voxel grids and 3D convolution. However, we use point clouds to store contact information for two reasons. (1) Contact locations and normals provide richer information for a single contact about the 3D object and yet are still simple enough to simulate. (2) The object is allowed to be anywhere in the workspace as long as the workspace center is occupied, resulting in a large workspace. Voxel grids are voluminous and scale cubically with the size of the workspace, taking up an unnecessarily large amount of memory. 

Our point cloud contains contact positions with contact normals as extra features. Compared to regular input data formats such as 3D voxels, point cloud data are unordered and can contain a  variable length of points. We choose PointNet++ as our network architecture to encode point clouds due to its permutation invariance of points and ability to handle variable input length. PointNet++ is a hierarchical extension of PointNet~\cite{qi2017pointnet} and uses PointNet as its basic building block for local pattern learning. 

In our encoder, there is a separate PointNet++ for both the discriminator and the explorer. Weights are not shared because they need to encode different aspects of the contact information, which leads to better performance. The two PointNet++'s have similar architecture but different inputs, as shown in Fig.~\ref{fig:overview}. For the discriminator, each point has a dimension of 3 (contact locations) + 3 (contact normals). For the explorer, the point cloud includes one extra point representing the current finger pose and each point has an extra bit of binary information indicating whether the point is the current finger pose or a regular contact pose. Without the important information about the current finger pose, the finger frequently gets stuck at workspace corners and is not able to navigate back toward the object.

The PointNet++ in our method has 2 set abstraction (SA) modules and 1 global SA module. Each SA module contains a sampling layer, a grouping layer, and a PointNet layer. In the sampling layer, a ratio of points is chosen as the centroid using the farthest point sampling (FPS) algorithm. In the grouping layer, points are grouped using the centroids with respect to a radius. In the PointNet layer, the local region of each group is abstracted by its centroid and local feature that encodes the centroid’s
neighborhood. The global SA module only has one PointNet layer that consumes the final abstracted centroids and their features.

Our 3D object representation is a feature vector of dimension 1024 produced by PointNet++. In the discriminator, the feature is passed through another MLP to generate the probability distribution over the 10 objects. The object with the highest probability is our prediction and the probability becomes the confidence. In the explorer, the feature is passed to the actor MLP to generate the action distribution and also the critic MLP to generate the state value for PPO training.

\subsection{Complete Exploration and Discrimination Architecture}

While the explorer and discriminator are distinct components, a key feature of TANDEM3D is that they are co-trained interleavingly, along with the PointNet++ encodings. The explorer is trained with reinforcement learning (RL) and a reward of 1 is given if its exploration is terminated by the discriminator, while the discriminator is trained with labeled data batches collected by the explorer. The effective encoding also allows us to train with a rich action space. As shown in the next section, the explorer learns to take advantage of this action space, combining translations and rotations of the sensor to quickly gather discriminative information. Overall, in the co-training process, each component affects the other, improves the other, co-evolves, and then converges. 

\section{Experiments}

In this section, we describe our experimental setup in both simulation and the real world\footnote{For real-world experiment videos or more information, please visit our project website at \url{https://jxu.ai/tandem3d}}. Our method is trained entirely in simulation and can be evaluated in simulation or with real hardware. We compare our method with a comprehensive set of baselines and then validate its performance on the real robot with a tactile sensor. 

\subsection{Experimental Setup}

We pick 10 objects from the YCB dataset as shown in Fig.~\ref{fig:teaser}. These objects are chosen to cover both convex and concave shapes and a variety of sizes intentionally. In simulation, we decompose each object into a set of convex parts using V-HACD~\cite{mamou2016volumetric} for collision checking.

The object is placed on a horizontal surface with its Z-axis pointing up. However, the rotation of the object around the vertical axis, as well as its translation along the axes of the horizontal plane are all unknown and randomized inside a 30cm by 30cm workspace. The only constraint is that some part of the object must occupy the center of the workspace so that the robot can make initial contact. In order to compensate for the calibration error in the real-world experiments, we also add a 2cm translation variance on the height of the objects during training in simulation. 

Real-world tactile sensors inevitably apply some level of force to the object before the contact is detected; for sensors with low sensitivity or light objects, the sensing act itself can thus induce movement. Handling such object movement during exploration is beyond the scope of this work, and we thus assume a sensor that can detect contact before causing movement. While such behavior is easy to simulate, in order to also achieve it in real life, we attach the object to the surface using Velcro. However, as can be seen in our accompanying video, some level of object tilt is inevitable after finger poking. Nevertheless, our algorithm is robust enough to such changes in object orientation. 

In real-world experiments, we use the DISCO~\cite{piacenza2020sensorized} finger (Fig.~\ref{fig:teaser}) to provide contact locations and normals. This multi-curved tactile finger has sensing abilities covering the hemisphere top and cylinder. We only keep contact information with a force magnitude larger than 0.5N and discard the rest because weaker contacts tend to have larger noise. We mount this finger on a UR5 robot arm to achieve 6DOF finger movement. In simulation, we use PyBullet to simulate a floating finger with the same tactile sensing capabilities.

Each episode is terminated if the number of steps surpasses 2,000 or the discriminator has a confidence value greater than the preset threshold of 0.98. When the episode is terminated, the prediction from the discriminator is compared to the true identity of the object to determine success. 




\subsection{Sensor Noise}
Being able to handle erroneous tactile readings without compromising efficiency and accuracy is critical for tactile-based applications. 
There are two types of sensor noise observed on our tactile finger.

\subsubsection{Contact Noise} This is the noise when a fake contact is reported when there is no contact. When it occurs, the output contact location can be anywhere on the finger surface. We quantify this noise as the percentage of time steps where the finger reports a non-existing contact. For our real sensors, we have empirically observed this to be approximately 0.1\%.

\subsubsection{Localization Noise} This is the error of the predicted contact location when a real contact happens. When contact occurs, both the contact locations and the contact normals deduced from them can be noisy. We quantify this as the distance in mm between the real contact location and the reported one. For our DISCO finger, the reported level for this noise is between 1 and 2 mm on average.

In our simulation experiments, we test all methods described below under two noise conditions. The first one is intended to emulate our real sensor: 0.1\% contact noise and 2 mm localization noise. The second condition attempts to emulate a better sensor, assuming future advances in sensing technology: 0.025\% contact noise and 0.5 mm localization noise. All real-world experiments are obviously subject to the noise level of the real tactile finger.

An important advantage of learning-based methods is that noise can be introduced during the training process, enabling the method to adapt. Thus, for both noise conditions described above, all the learning-based methods discussed below are trained under the respective noise conditions.

\begin{table*}
    \caption{Comparative performance of various methods in simulation. Methods are trained and evaluated under two noise conditions. For each method, we present the number of actions taken (\#Actions) and the number of points explored  (\#Points) at termination, as well as the success rate in identifying the correct object (Success Rate). Mean and standard deviation over 1,000 trials are shown. A detailed description of each method can be found in Sec.~\ref{sec:methods}.}
    \label{tab:results}
    \centering
    \begin{tabular}{c|ccc|ccc}
    \toprule
    \multirow{2}{*}{\textbf{Method}} & \multicolumn{3}{c|}{\textit{0.025\% Contact Noise, 0.5mm Localization Noise}} & \multicolumn{3}{c}{\textit{0.1\% Contact Noise, 2mm Localization Noise}} \\
    & \#\textbf{Actions} & \textbf{\#Points} & \textbf{Success Rate} & \#\textbf{Actions} & \textbf{\#Points} & \textbf{Success Rate} \\
    \midrule
    Not-go-back &$360.6\pm315.9$ & $4.666\pm2.847$ & $0.68$ & $294.2\pm289.9$ & $4.677\pm2.579$ & $0.52$  \\
    Info-gain & $678.2\pm318.1$ & $2.486\pm1.725$ & $0.29$ & $536.7\pm336.2$ & $2.988\pm1.875$ & $0.23$ \\   
    Edge-follower & $57.80\pm133.4$ & $12.21\pm27.10$ & $0.89$ & $59.22\pm142.2$ & $10.43\pm20.56$ & $0.83$ \\
    Edge-ICP & $85.30\pm126.7$ & $20.92\pm16.75$ & $0.26$ & $93.77\pm162.2$ & $19.84\pm14.52$ & $0.26$ \\
    PPO-ICP & $117.4\pm141.8$ & $16.95\pm11.09$ & $0.13$  & $109.0\pm126.4$ & $16.29\pm9.550$ & $0.14$ \\
    All-in-one & $118.3\pm268.8$ & $6.613\pm2.277$ & $0.13$ & $94.65\pm162.9$ & $6.669\pm2.114$ & $0.12$  \\  
    TANDEM3D (ours) & $45.73\pm96.88$ & $9.416\pm9.831$ & $0.98$ & $43.02\pm71.08$ & $9.806\pm10.61$ & $0.93$ \\
    \bottomrule
    \end{tabular}
\end{table*}

\subsection{Baselines}
\label{sec:methods}


We compare our approach to a variety of learning-based and heuristic-based baselines explained below. The metrics that we are most interested in are the number of actions and the success rate in accurately identifying the objects.

\subsubsection{Not-go-back} This exploration policy picks a random move that leads to an unexplored finger pose at each step. A discriminator is trained with this exploration policy.

\subsubsection{Info-gain} This method uses the info-gain heuristic. It chooses an action that provides the most salient information and leads to an unexplored finger pose. At the current time step $t$, let $\mathbf{p}$ denote the probability distribution over 10 objects produced by the discriminator. Let $\mathbf{p_{c}}$ and $\mathbf{p_{n}}$ denote the new probability distributions after applying a particular action when a new contact happens or not respectively. Clearly, $\mathbf{p_n} = \mathbf{p}$ because the predicted distribution does not change without new contacts. When computing $\mathbf{p_c}$, we assume the new contact location is on the top of the fingertip hemisphere. The action $a_t$ is chosen by: 
\begin{align*}
    a_t &= \argmax_{a \in \mathcal{A}} \: \bigg\{ \mathcal{H} (\mathbf{p}) - \left(\frac{1}{2} \mathcal{H} (\mathbf{p_c}) + \frac{1}{2} \mathcal{H} (\mathbf{p_n})\right) \bigg\} \\
    &= \argmax_{a \in \mathcal{A}} \: \left\{ \mathcal{H} (\mathbf{p}) - \mathcal{H} (\mathbf{p_c}) \right\}
\end{align*} 
where $\mathcal{H}$ denotes the entropy of a probability distribution. It uses entropy as a measure of uncertainty and picks the action that reduces the most uncertainty. A discriminator is trained along with the explorer. 

\subsubsection{Edge-follower} This method uses the contour-following heuristic. However, contour-following on a 3D object is not well-defined. At any point on a 3D surface, there is an infinite number of candidate edges to follow. In this implementation, the finger starts following a horizontal edge parallel to the workspace plane at the initial contact height. The finger angle is adjusted to avoid collision between the finger bottom and the object. We also implement a variant that follows the vertical edge parallel to the workspace XZ-plane but it performs much worse than our horizontal version. The reason is that depending on the initial position of the object and the initial contact location, the intersection of the object with the vertical edge plane can vary, potentially leading to a very small contour. Such a contour contains little information and can result in recognition failure. \textit{Edge-follower} is the only baseline that is not trained with contact noise. The performance drops significantly when applying contact noise during training because \textit{Edge-follower} can get trapped at fake contact locations and starts circling that location. In such a case, it can not continue exploring and the discriminator becomes unnecessarily cautious but the exploration policy is not able to increase its confidence.



\subsubsection{Edge-ICP} This method uses the same edge-following exploration heuristic but instead of training a discriminator, it uses the Iterative Closest Point (ICP) algorithm for recognition. We randomly sample 1,000 points on object mesh surfaces that are approximately evenly spaced as the ground truth point clouds. The discriminator uses ICP to match the point cloud with 36 different initial orientations linearly spaced between [0\degree, 360\degree]. For each object, the minimum error among all orientations represents the overall matching error. If the error is smaller than a threshold, it is marked as a match. Our threshold decays linearly from $5e^{-3}$ with $5e^{-5}$ step size. The output probability distribution assigns equal probabilities to the matched objects and zeroes to not-matched ones. This method requires no training.

\subsubsection{PPO-ICP} This method trains a PPO explorer as in ours with the ICP discriminator as in \textit{Edge-ICP}.

\subsubsection{All-in-one} This method does not separate explorer and discriminator. It has the same structure as the PPO explorer in our method but the action space is expanded to 24 actions. The first 12 actions indicate a move and the remaining 10 actions indicate a prediction. If the explorer picks a prediction instead of a move, the episode is terminated. A reward of 1 is given when an accurate prediction is made. 


\subsubsection{TANDEM3D} This is our proposed method. Training takes 120 hours for the high-noise condition on an Nvidia RTX 3090 GPU and an Intel i9-12900K CPU. 

\begin{figure}[]
\centering
\includegraphics[width=\linewidth]{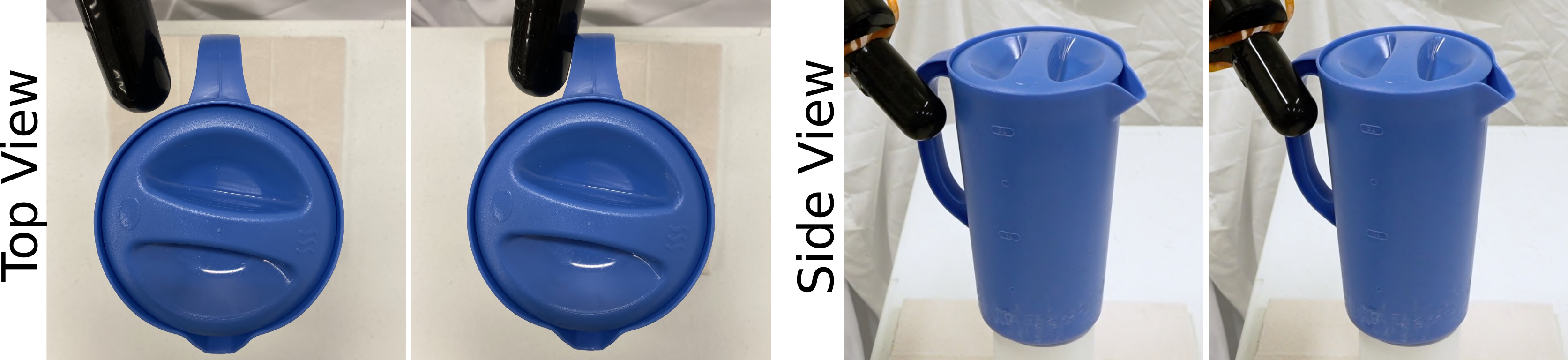}
\caption{\textit{TANDEM3D} angle adjustment move. Our tactile finger with all-around sensing coverage collects a critical contact point on the pitcher handle by adjusting the angle.}
\label{fig:swing}
\end{figure}

\begin{figure}[]
\centering
\includegraphics[width=\linewidth]{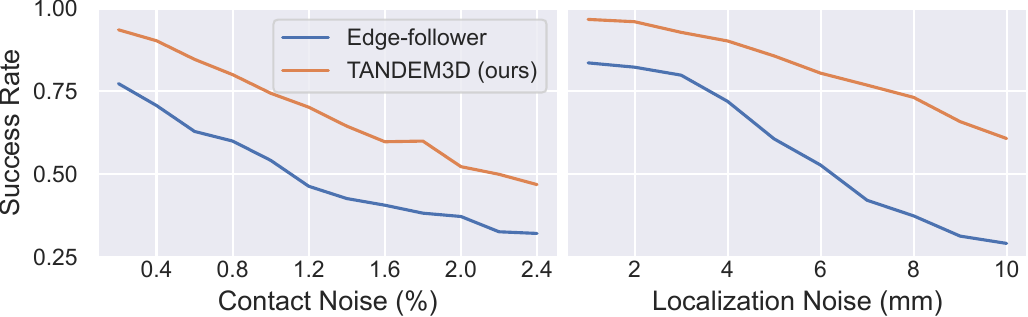}
\caption{Robustness and generalization to larger sensor noise. Despite being trained on lower noise levels, \textit{TANDEM3D} retains a high success rate as the contact noise increases to 2.4\% or the localization noise increases to 10mm, while \textit{Edge-follower}'s performance worsens more significantly.}
\label{fig:noise}
\vspace{-0.15in}
\end{figure}

\begin{figure*}[]
\centering
\includegraphics[width=0.98\textwidth]{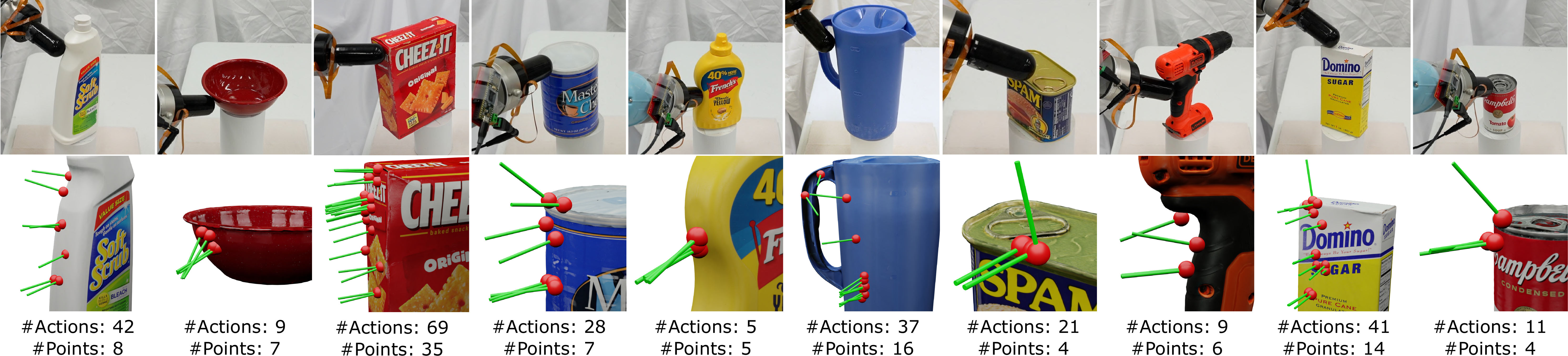}
\vspace{-0.05in}
\caption{10 successful examples of our method in real robot experiments. The top row shows the object poses and the final finger poses. The medium row shows the collected contact positions and normals at termination. The last row shows the results for each trial. Our method reacts to previous observations along the trajectory and gradually moves to the most distinguishable areas for efficient recognition. Each example is accompanied by a real-world video on our project website at \url{https://jxu.ai/tandem3d}.}
\label{fig:real}
\vspace{-0.24in}
\end{figure*}




\subsection{Performance Analysis and Discussion}

We evaluate all methods with 1,000 simulated trials and the results are shown in Table~\ref{tab:results}. 

\textit{TANDEM3D} outperforms all other baselines. It learns an exploration behavior that combines following approximate vertical edges (either upwards or downwards) and swinging finger by adjusting angles, as illustrated in the real-world examples in Fig.~\ref{fig:real}. The angle adjustment enables the robot to collect contacts diverging from the vertical trajectory quickly which can provide critical information about object geometry. The swing motion also suits very well with our multi-curved tactile finger which has large sensing coverage. As shown in Fig.~\ref{fig:swing}, with the small angle adjustment, the collision between the finger cylinder and pitcher handle produces a contact position far from the current fingertip location. With such a discriminative contact, the prediction quickly converges to the correct object.

\textit{Edge-follower} has the closest performance to ours. However, without the flexible swing motions, it has to follow a large portion of the edge in order to gather the same amount of information. In addition, \textit{Edge-follower} is much more sensitive to both contact noise and localization noise. With contact noise, the finger can get stuck in following a fake contact. With larger localization noise, the points diverge from the contour trajectory which \textit{Edge-follower} overfits to, leading to a poor success rate.

To further compare the robustness and generalization of \textit{TANDEM3D} and \textit{Edge-follower} to larger sensor noise than what they are trained on, we evaluate both methods with contact noise up to 2.4\% and localization noise up to 10mm. Note that, in this condition, noise levels are only increased for testing, and not for training. As shown in Fig.~\ref{fig:noise}, the success rate of \textit{Edge-follower} deteriorates more dramatically as sensor noise increases beyond the levels seen in training.

Despite being proven to be a useful exploration heuristic in many previous works, \textit{Info-gain} performs surprisingly badly. We attribute its unsatisfying performance to the fact that we are training a discriminator along with it from scratch. Unlike other explorers, the \textit{Info-gain} explorer relies on the output from the discriminator at each time step. It needs a good discriminator to start with. We also notice that the finger tends to leave the object instead of approaching it during exploration. We think that the \textit{Info-gain} explorer attempts to make a contact that is far from the center as those points tend to provide the most salient information for discrimination. 

The methods with classic ICP recognition also demonstrate low accuracy due to the limited number of points in the partial point clouds. Taking \textit{Edge-ICP} as an example, even with a complete horizontal edge-following trajectory, the point clouds gathered for each object are mostly circles and ellipses with a few points. It is difficult for the ICP algorithm to match such simple shapes to a full 3D point cloud with 1,000 points. The \textit{All-in-one} baseline does not have a dedicated discriminator that can be co-trained with batches of labeled samples collected by the explorer, leading to inefficient training and low performance.

\subsection{Real-world Performance}

We test the performance of TANDEM3D in real-world experiments. For each object, we run 2 trials with the object randomly placed on the support in its upright orientation. The results are shown in Table~\ref{tab:real}. Our method still achieves a high success rate (although slightly lower than in simulation), thanks to its ability to generalize in the presence of noise. The sim-to-real gap when transferring our trained models to the real robot is largely due to workspace calibration, change of object pose under contact, mesh discrepancy caused by convex decomposition, etc. For example, real object 2 (cracker box) has distorted faces which introduce a large mismatch.  We select one successful example per object from our real-world experiments and visualize the object poses, final finger poses, collected contact positions, and contact normals before a final prediction is made in Fig.~\ref{fig:real}.

\begin{table}[h!]
    \caption{Real robot experiment results (mean and standard deviation over 20 trials).\vspace{-0.1in}}
    \label{tab:real}
    \centering
    \begin{tabular}{c|ccc}
    \toprule
    \textbf{Method} & \#\textbf{Actions} & \textbf{\#Points} &  \textbf{Success Rate} \\
    \midrule
    TANDEM3D & 38.21 $\pm$ 25.16 & 10.31 $\pm$ 10.93 & 0.85 (17/20) \\  
    \bottomrule
    \end{tabular}
\end{table}

\vspace{-0.05in}
As we can see, \textit{TANDEM3D} reacts to previous observations and takes moves to the most discriminative areas. For example, on object 0 (bleach cleanser), the finger moves to the back of the object and contacts the recess, which is a distinguishable geometry special to object 0. On object 5 (pitcher base), the finger moves up towards the pitcher handle and then makes a decision by making a small angle adjustment to contact the handle using the side of the finger. On object 7 (power drill), the finger swings up to contact from beneath the drill and immediately makes the correct prediction. Please see our accompanying video for a better demonstration of these behaviors.

\section{Conclusion}

We present TANDEM3D, a new architecture to co-train exploration and decision making for 3D object tactile recognition. Our method enables 6DOF movements of the tactile sensor and is able to discover discriminative points through small angle adjustments, taking advantage of a tactile finger with all-around sensing coverage. Our method is based on an encoder that builds object representation from the contact positions and surface normals acquired via tactile sensing. We train our method entirely in simulation and zero-shot transfer to the real robot. Compared to state-of-the-art alternatives, our method can correctly identify 3D objects with fewer actions and a higher success rate.  It also demonstrates better generalization ability to different types and amounts of sensor noise. Future directions include extending to category-level object classification with many object instances per category and studying multi-finger exploration policies.








\bibliographystyle{IEEEtran}
\bibliography{reference}

\end{document}